\pdfoutput=1

\documentclass[11pt]{article}

\usepackage{ACL2023}

\usepackage{times}
\usepackage{latexsym}
\usepackage{xspace}
\usepackage{multirow}
\usepackage{booktabs}
\usepackage{graphicx}
\usepackage{amsmath}
\usepackage{dsfont}
\usepackage{scrextend}
\usepackage{subcaption}
\usepackage{paralist}
\usepackage{titlesec}

\usepackage[T1]{fontenc}

\usepackage[utf8]{inputenc}

\usepackage{microtype}

\usepackage{inconsolata}
\newcommand{\ours}{\textsc{Ear}\xspace}
\newcommand{\GAR}{\textsc{Gar}\xspace}
\newcommand{\mytilde}{{\raise.17ex\hbox{$\scriptstyle\sim$}}}

\title{Expand, Rerank, and Retrieve: \\Query Reranking for Open-Domain Question Answering}

\author{Yung-Sung Chuang$^\dagger$ \quad Wei Fang$^\dagger$ \quad Shang-Wen Li$^\ddagger$ \quad Wen-tau Yih$^\ddagger$ \quad James Glass$^\dagger$ \\
  Massachusetts Institute of Technology$^\dagger$ \quad
  Meta AI$^\ddagger$ \\
  \texttt{yungsung@mit.edu} \\
    }

\begin{document}
\maketitle
\begin{abstract}
We propose \ours, a query \textbf{E}xpansion \textbf{A}nd \textbf{R}eranking approach for improving passage retrieval, with the application to open-domain question answering. \ours first applies a query expansion model to generate a diverse set of queries, and then uses a \textit{query} reranker to select the ones that could lead to better retrieval results. Motivated by the observation that the best query expansion often is not picked by greedy decoding, \ours trains its reranker to predict the rank orders of the gold passages when issuing the expanded queries to a given retriever. By connecting better the query expansion model and retriever, \ours significantly enhances a traditional sparse retrieval method, BM25. Empirically, \ours improves top-5/20 accuracy by 3-8 and 5-10 points in in-domain and out-of-domain settings, respectively, when compared to a vanilla query expansion model, \GAR, and a dense retrieval model, DPR.\footnote{Source code: \url{https://github.com/voidism/EAR}.}

\end{abstract}

\section{Introduction}

Open-domain question answering (QA)~\cite{chen-yih-2020-open}, a task of answering a wide range of factoid questions of diversified domains, is often used to benchmark machine intelligence~\cite{kwiatkowski-etal-2019-natural} and has a direct application to fulfilling user's information need~\cite{voorhees1999trec}. To provide faithful answers with provenance, and to easily update knowledge from new documents, passage \emph{retrieval}, which finds relevant text chunks to given questions, is critical to the success of a QA system.
Retrieval in early open-domain QA systems~\cite{chen-etal-2017-reading} is typically based on term-matching methods, such as BM25~\citep{robertson2009probabilistic} or TF-IDF~\cite{Salton75}. 
Such methods are sometimes called \emph{sparse} retrievers, as they represent documents and queries with high-dimensional sparse vectors, and can efficiently match keywords with an inverted index and find relevant passages.
Despite their simplicity, sparse retrievers are limited by their inability to perform semantic matching for relevant passages that have low lexical overlap with the query.
Lately, dense retrievers~\citep{karpukhin-etal-2020-dense}, which represent documents and queries with dense, continuous semantic vectors, have been adopted by modern QA systems. Dense retrievers usually outperform their sparse counterparts, especially when there exists enough in-domain training data.

However, dense retrievers have certain weaknesses compared to sparse ones, including: 1) being computationally expensive in training and inference, 2) potential information loss when compressing long passages into fixed-dimensional vectors~\citep{luan-etal-2021-sparse}, which makes it hard to match rare entities exactly~\citep{sciavolino-etal-2021-simple}, and 3) difficulty in generalizing to new domains~\citep{reddy2021towards}. As a result, dense retrievers and sparse ones are usually complementary to each other and can be combined to boost performance. 
Recent studies on query expansion, such as \GAR~\citep{mao-etal-2021-generation}, have attempted to improve sparse retrievers by adding relevant contexts to the query using pre-trained language models (PLMs), which has been shown effective in closing the gap between sparse and dense retrievers.

\begin{figure*}[ht!]
    \centering
    \includegraphics[width=1.0\linewidth]{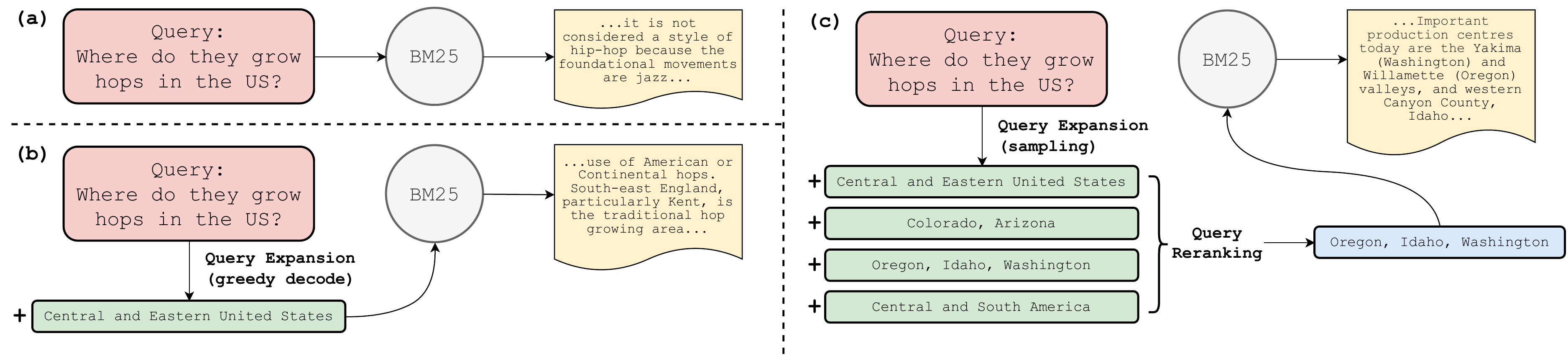}
    \caption{(a) Standard BM25 pipeline (b) Generation-Augmented Retrieval (\GAR) with BM25 (c) Our proposed Expand and Rerank (\ours) pipeline.}
    \label{fig:ear}
\end{figure*}

In this paper, we introduce a novel query \textbf{E}xpansion \textbf{A}nd \textbf{R}eranking approach, \ours, which enhances generative query expansion with \textbf{query reranking}. 
\ours first generates a diverse set of expanded queries with query expansion models, and then trains a query reranker to estimate the \emph{quality} of these queries by directly predicting the rank order of a gold passage, when issuing these queries to a given retriever, such as BM25.
At inference time, \ours selects the most promising query expansion as predicted by the query reranker and issues it to the same retriever to find relevant documents. 
\ours is motivated by a simple observation---while the greedy decoding output of a query expansion model, such as~\GAR, could be suboptimal, some randomly sampled query expansions achieve superior performance with BM25 (see Section~\ref{sec:prelim}). 
\ours better connects the query expansion model and the underlying retrieval method, and thus can select a more suitable query. 

We empirically evaluated \ours in both \emph{in-domain} and \emph{cross-domain} settings. Our \emph{in-domain} experimental results on Natural Questions and TriviaQA show that \ours significantly improves the top-5/20 accuracy by 3-8 points. 
For the \emph{cross-domain} setting, while the query expansion model suffers from substantial performance degradation when applied to new domains, \ours seems to be more domain-agnostic, and can still find useful queries from a diverse set of query expansions, which leads to a significant improvement over \GAR and DPR by 5-10 points for top-5/20 accuracy.

\begin{table}[t!]
    \centering
    \small
    \resizebox{0.48\textwidth}{!}{%
    \begin{tabular}{l|cccc}
    \toprule
       \bf Model & \bf Top-1 & \bf Top-5 & \bf Top-20 & \bf Top-100 \\
    \midrule
        1) BM25 & 22.1 & 43.8 & 62.9 & 78.3 \\
        2) DPR & 43.0 & 66.4 & 78.5 & 85.0 \\
        3) \GAR (greedy) & 37.0 & 60.8 & 73.9 & 84.7 \\
        4) \GAR (beam=10) & 38.6 & 61.6 & 75.2 & 84.8 \\
        5) \emph{\GAR best query} & 68.8 & 81.9 & 88.1 & 92.0 \\
        6) \emph{\GAR concat} & 39.5 & 60.3 & 72.7 & 83.6 \\
    \bottomrule
    \end{tabular}
    }
    \caption{
        The potential top-k retrieval accuracy that can be achieved by query reranking on Natural Questions. \GAR uses greedy-decoded/beam-searched queries; \emph{\GAR best query} randomly samples 50 queries and picks the oracle one with the best retrieval scores; \emph{\GAR concat} simply concatenates all 50 queries as a single long query.
    }
    \label{tab:oracle}
\end{table}

\noindent Our contributions can be summarized as follows:
\begin{compactitem}
    \item We proposed \ours to select the best query from a diverse set of query expansions, by predicting which query can achieve the best BM25 result. This improves the connection of query expansion models and BM25, resulting in enhanced performance that surpasses DPR.
    \item \ours not only performs well on in-domain data, but also shows strong generalization abilities on out-of-domain data, outperforming \GAR and DPR by a large margin.
    \item End-to-end evaluation with a generative reader demonstrates the benefits of \ours in improving the exact match score.
    \item Lastly, we show that the improvements provided by \ours and passage reranking are complementary, allowing for effective aggregation of performance gains from both methods.
\end{compactitem}

\section{Background}
\label{sec:primary}

\subsection{Generation-Augmented Retrieval}
Generation-Augmented Retrieval (\GAR)~\citep{mao-etal-2021-generation} aims to enhance sparse retrievers by query expansion with text generation from PLMs. Given the initial query, \GAR generates relevant contexts including \emph{the answer, answer sentence}, and \emph{title of answer passages}, and then concatenates them to the initial query before performing retrieval with BM25.
\GAR achieves decent performance close to that of DPR while using the lightweight BM25 retriever. 
However, a limitation is that \GAR is not aware of the existence of BM25, potentially generating suboptimal queries for retrieval. Additionally, \GAR is only trained on in-domain data, limiting their ability to transfer to out-of-domain data.

\subsection{Preliminary Experiments}
\label{sec:prelim}
Let us first take a look at some preliminary experimental results to better understand the motivation of this paper. In Table~\ref{tab:oracle}, we present the top-k retrieval results on Natural Questions~\citep{kwiatkowski-etal-2019-natural} for BM25, DPR, and \GAR (greedy/beam search) in rows 1-4. To investigate the potential of \GAR, we randomly sampled 50 query expansions from \GAR, ran BM25 separately for these queries, and chose the best one by looking at the BM25 results, which requires ground truth labels. The resulting scores are shown in row 5 (\emph{\GAR best query}).

From the results, we see that \emph{\GAR best query} can lead to a significant improvement of up to 20 points compared to DPR. Since we do not have access to labels for selecting the best query in reality, a naive solution is to concatenate all 50 query expansions together as a single, long query, which will definitely include high-quality expansions if they exist. However, as shown in row 6, the performance of \GAR \emph{concat} is even worse than that of \GAR alone with greedy decoding outputs. This indicates that the single long query may include too much distracting information, negatively impacting the performance of the BM25 retriever.

From these preliminary results, we reach two conclusions: 1) \GAR does have the ability to generate very useful query expansions; 2) however, the useful query expansions may not always be included in the \GAR greedy decoding outputs. It is non-trivial to extract these useful query expansions from \GAR. Motivated by these findings, we leverage a query reranker to estimate if a query will be beneficial to BM25 retrieval results, so as to unlock the potential of \GAR and sparse retrievers.

\section{Proposed Method}

We illustrate our proposed method, \ours, in Figure~\ref{fig:ear}, along with a comparison with the BM25 and \GAR pipelines. Given the original query $q$, \ours first generates a set of query expansions $E = \{e_1, e_2, ..., e_n\}$ using random sampling. We believe that among these $n$ queries, some may achieve very good retrieval performance. Thus, we train a reranker model $\mathcal{M}$ to re-score all the queries. Here we propose two kinds of rerankers: 1) retrieval-independent (RI) reranker, and 2) retrieval-dependent (RD) reranker. Both rerankers can estimate the quality of a query expansion without using information from answer annotations.

\subsection{Retrieval-Independent (RI) Reranker}
\label{sec:ri}

The inputs to the RI reranker are quite simple: $(q, e_i)$, which consists of the original query $q$ and one of the query expansions $e_i$. When training this reranker, we first obtain the minimum answer passage ranking among all retrieved passages for each query, when issued to a BM25 retriever. We denote this minimum answer passage ranking as $R = \{r_1, r_2, ..., r_n\}$, which corresponds to each of the expanded queries $\{(q, e_1), (q, e_2), ..., (q, e_n)\}$. 

To clarify the concept, let us consider an example with two query expansions, $e_1$ and $e_2$. Say the expanded query $(q, e_1)$ retrieves the answer passage as the top result (first position), we assign $r_1 = 1$. Similarly, we assign $r_2 = 15$ if the expanded query $(q, e_2)$ retrieves the answer passage in the 15th position. In this case, we conclude that $e_1$ is a better query expansion than $e_2$ since its corresponding ranking value, $r_1$, is lower than $r_2$.

$r_i$ can be seen as the \emph{score} that can be obtained by the query of $(q, e_i)$, with smaller $r_i$ corresponding to better quality of $(q, e_i)$. 
We now train a scoring model to estimate the rank $r_i$ for given inputs $(q, e_i)$. However, considering that the scoring model will be used as a reranker, we only need to ensure the model's \emph{relative} accuracy of estimating $r_i$, rather than its \emph{absolute} value. Thus, we employ a ``contrastive'' loss rather than a ``regression'' loss, which is inspired by the contrastive method in summarization re-scoring~\citep{liu-liu-2021-simcls}.

For all pairs of query expansions $(e_i, e_j)$ such that $r_i < r_j$ (which means $e_i$ is a better expansion than $e_j$), the ranking loss is calculated as follows:
\vspace{-10pt}
\begin{multline*}
\mathcal{L}_{\text{Rank}}= \\
\sum_{\substack{i,j \in [1,n] \\ r_{i}<r_{j}}} \max(0, \mathcal{M}(q, e_{i})-\mathcal{M}(q, e_{j})+(r_j - r_i)\cdot\alpha)
\vspace{-10pt}
\label{eq:rank}
\end{multline*}
Here, $\mathcal{M}$ is a model that estimates the rank $r_i$ for a given query expansion $e_i$. Instead of predicting the absolute rank of $r_i$, the model $\mathcal{M}$ is trained to predict the difference between $r_i$ and $r_j$ for each pair of expansion $(e_i, e_j)$.

The ranking loss $\mathcal{L}_{\text{Rank}}$ forces the model to estimate a lower rank for $e_i$ and a higher rank for $e_j$, such that the difference between $\mathcal{M}(q, e_{i})$ and $\mathcal{M}(q, e_{j})$ is greater than the threshold $(r_{j} - r_{i}) \cdot \alpha$, where $\alpha$ is a scalar. If some of the expansions do not retrieve the answer passages within the top-$k$ results (e.g. within the top-100 results), we assign a constant value, \texttt{MAX\_RANK}, to these expansions.

\subsection{Retrieval-Dependent (RD) Reranker}

The input to the RI Reranker only contains the original query $q$ and the expansion $e_i$, which may not be sufficient to distinguish good expansions from bad expansions. For example, in Figure~\ref{fig:ear} (c), for the original query \textit{Where do they grow hops in the US?}, it is easy to tell that \textit{Central and South America} is a bad expansion because the US is not in Central and South America. However, for these two expansions: 1) Colorado, Arizona 2) Oregon, Idaho, Washington, it is very hard to tell which one is better without any external knowledge. To alleviate this problem, we propose the Retrieval-Dependent (RD) Reranker, which is able to see the top-1 passages $D = \{d_1, d_2, ..., d_n\}$ retrieved by each query expansion.~\footnote{For RD reranker, we need additional computational costs to retrieve the top-1 passage. However, this process can be efficiently parallelized for all queries using a lightweight BM25 retriever, so the required time is not excessive. We will discuss the latency of \ours further in Section~\ref{sec:latency}.} The inputs of RD reranker will contain the original query $q$, the query expansions $e_i$, and the top-1 passage $d_i$. We train RD reranker with the same ranking loss $\mathcal{L}_{\text{Rank}}$, but replace the model with $\mathcal{M}(q, e_{i}, d_{i})$.

\begin{table*}[h!]
\centering
\small
\begin{tabular}{lcccccc} 
\toprule
\multirow{2}{*}{\bf Model}
& \multicolumn{3}{c}{\bf Natural Questions~}  
& \multicolumn{3}{c}{\bf TriviaQA}       \\ 
\cmidrule(lr){2-4} \cmidrule(lr){5-7}
\multicolumn{1}{c}{} & \bf Top-5 & \bf Top-20 & \bf Top-100
& \bf Top-5 & \bf Top-20 & \bf Top-100   \\ 
\midrule
\multicolumn{7}{c}{\textit{Dense Retrieval}}\\
\midrule
DPR & 68.3 & 80.1 & 86.1 & 72.7 & 80.2 & 84.8 \\
\midrule
\multicolumn{7}{c}{\textit{Lexical Retrieval}}\\
\midrule
BM25 & 43.8 & 62.9 & 78.3 & 67.7 &
77.3 & 83.9  \\
\GAR & 60.8 & 73.9 & 84.7 & 71.8 & 79.5 & 85.3 \\
SEAL & 61.3 & 76.2 & 86.3 & - &  - & - \\
\citet{liu2022query} & 63.9 & 76.8 & \bf 86.7 & 72.3 & 80.1 & 85.8  \\
\midrule
\ours-RI & 63.2 & 76.4 & 85.9 & 73.4 & 80.8 & 85.9 \\
\ours-RD & \bf 69.3 & \bf 78.6 & 86.5 & \bf 77.6 & \bf 82.1 & \bf 86.4 \\
\midrule
\GAR \emph{best query} & \it 81.9 & \it 88.1 & \it 92.0 & \it 85.0 & \it 88.1 & \it 90.1 \\

\midrule
\multicolumn{7}{c}{\textit{Fusion (Dense + Lexical) Retrieval}}\\
\midrule
BM25 + DPR & 69.7 & 81.2 & 88.2 &71.5 & 79.7 & 85.0 \\
\GAR + DPR & 72.3 & \bf 83.1 & 88.9 & 75.7 & 82.2 & 86.3 \\
\citet{liu2022query} + DPR & 72.7 & 83.0 & 89.1 & 76.1 & 82.5 & 86.4 \\
\midrule
\ours-RI + DPR & 71.1 & 82.5 & 89.1 & 76.4 & 83.0 & 87.0 \\
\ours-RD + DPR & \bf 74.2 & \bf 83.1 & \bf 89.3 & \bf 79.0 & \bf 83.7 & \bf 87.3 \\
\bottomrule
\end{tabular}
\caption{Top-k retrieval accuracy (\%) on the NQ and TriviaQA test sets. Numbers for prior work are cited from \citet{liu2022query}.}
\label{tab:main_result}
\end{table*}
\subsection{Training Examples Construction}
\label{sec:construct}

To construct training examples, we generate diverse query expansions, run BM25 retrieval on them, and train the rerankers based on the results. However, using the \GAR generators directly may not yield diverse sequences and limit the rerankers' learning, since the \GAR generators are trained with supervision and may have already overfit on the training set, which would lead to almost identical generation samples. To address this, we propose two alternatives: 1) Split the training set into $K$ subsets, train $K$ different GAR generators on ($K$-1) subsets and randomly sample from the remaining subset; and 2) Use a large language model (LLM) such as T0~\citep{sanh2021multitask} to randomly sample query expansions directly without fine-tuning. Both options performed equally well in our experiments and will be further discussed in Section~\ref{sec:t0}.

\section{Experiments}

\subsection{Data}

For in-domain experiments, we use two public datasets for training and evaluation: Natural Questions (NQ)~\citep{kwiatkowski-etal-2019-natural} and TriviaQA~\citep{joshi-etal-2017-triviaqa}. For out-of-domain (cross-dataset) experiments, we directly evaluate our in-domain models on three additional public datasets without using their training sets: WebQuestions (WebQ)~\citep{berant-etal-2013-semantic}, CuratedTREC (TREC)~\citep{baudivs2015modeling}, and EntityQuestions (EntityQs)~\citep{sciavolino-etal-2021-simple}. (See dataset statistics in Appendix~\ref{app:data}.) All experiments are performed with Wikipedia passages used in DPR~\citep{karpukhin-etal-2020-dense}, consisting of 21M 100-word passages from the English Wikipedia dump of Dec. 20, 2018~\citep{lee-etal-2019-latent}. 

\subsection{Setup}

\paragraph{Model}
For sparse retrieval, we use Pyserini~\citep{Lin_etal_SIGIR2021_Pyserini} for BM25 with its default parameters. For query rerankers, we use the DeBERTa V3 base~\citep{he2021debertav3} model from Huggingface Transformers~\citep{wolf-etal-2020-transformers}. For RI reranker, the input format is: \texttt{[CLS] <question> ? <expansion> [SEP]}; for RD reranker, the input format is \texttt{[CLS] <question> ? <expansion> [SEP] <top-1 retrieved passage> [SEP]}.
Training details can be found in Appendix~\ref{app:training}. 

\paragraph{Context Generator}
At training time, we use T0-3B~\citep{sanh2021multitask} to randomly sample 50 query expansions per question, as we mentioned in Section~\ref{sec:construct}. We add a short prompt, \emph{To answer this question, we need to know}, to the end of the original question, and let T0-3B complete the sentence. During inference, we still use the \GAR generators to randomly sample 50 query expansions per question on the testing set, since the examples are not seen during \GAR training and the generations are diverse enough. To speed up the inference process, we de-duplicate the query expansions that appear more than once. 
The average number of query expansions we use is 25 for Natural Questions and 34 for TriviaQA, respectively.
\subsection{Baselines}
We compare \ours with 1) DPR~\citep{karpukhin-etal-2020-dense}: a standard BERT-based dense retriever; 
2) BM25~\citep{robertson2009probabilistic}: a standard sparse retriever based on term matching; 
3) \GAR~\citep{mao-etal-2021-generation}: generation-augmented retrieval with BM25; 
4) \citet{liu2022query}: a concurrent work that uses a \GAR-like generative model to perform beam search decoding, followed by filtering to obtain multiple expanded queries for performing multiple retrievals with BM25, and then fusion of the results; 
and 5) SEAL~\citep{bevilacqua2022autoregressive}: an autoregressive search engine, proposing constrained decoding with the FM-index data structure that enables autoregressive models to retrieve passages.

\subsection{Result: In-Domain Dataset}
We first train and evaluate \ours on NQ and TriviaQA. In Table~\ref{tab:main_result}, we see that both \ours-RI and \ours-RD improve the performance of \GAR significantly. \ours-RI improves the top-5/20/100 accuracy of \GAR by 1-2 points, while \ours-RD improves the top-5 accuracy of \GAR by 6-8 points, and the top-20 accuracy by 3-5 points on both datasets. Moreover, \ours-RD is significantly better than DPR except for the top-20 accuracy on NQ. These results show that it is possible for BM25 to beat dense retrieval with the help of an optimized process to generate high-quality query expansions. Additional qualitative studies in Appendix~\ref{app:qual} provide further insight into how \ours works. We also report the results of \emph{the best query from \GAR}, which presents the potential performance upper bound that could be achieved by query reranking. It suggests that there is still room for \ours to improve if mechanisms for more effective query selection are developed. At the bottom of Table~\ref{tab:main_result}, we present the fusion retrieval results of combining \ours and DPR. \ours-RD+DPR outperforms the fusion results of BM25/\GAR/\citet{liu2022query}, showing the complementarity between \ours-RD and DPR.

\begin{table*}[h!]
\centering
\small
\begin{tabular}{lccccccccc}
\toprule
\multirow{2.6}{55pt}{\textbf{Model}} & 
\multicolumn{3}{c}{\textbf{WebQuestions}} & \multicolumn{3}{c}{\textbf{TREC}} & \multicolumn{3}{c}{\textbf{EntityQuestions}}
\\ \cmidrule(lr){2-4} \cmidrule(lr){5-7} \cmidrule(lr){8-10} 
& \textbf{Top-5} & \textbf{Top-20} & \textbf{Top-100} & \textbf{Top-5} & \textbf{Top-20} & \textbf{Top-100}& \textbf{Top-5} & \textbf{Top-20} & \textbf{Top-100}  \\
\midrule
BM25 & 41.8 & 62.4 & 75.5 & 64.3 & 80.7 & 89.9 & 60.6 & 70.8 & 79.2 \\

\midrule
& \multicolumn{9}{c}{\textit{In-Domain Supervised}}  \\
\midrule
DPR$^\dagger$ & 62.8 & 74.3 & 82.2 & 66.6 & 81.7 & 89.9 & - & - & -  \\ 

\midrule
& \multicolumn{9}{c}{\textit{Transfer from NQ}}  \\
\midrule
DPR$^\dagger$ & 52.7 & 68.8 & 78.3 & 74.1 & 85.9 & 92.1 & 38.1 & 49.7 & 63.2 \\ 
\GAR & 50.0 & 66.0 & 79.0 & 70.9 & 83.9 & 92.4 & 59.7 & 71.0 & 79.8 \\
\ours-RI & 53.7 & 69.6 & 81.3 & 73.5 & 85.9 & 92.9 & 62.7 & 73.3 & 81.4 \\
\ours-RD & \bf 59.5 & \bf 70.8 & \bf 81.3 & \bf 80.0 & \bf 88.9 & \bf 93.7 & \bf 65.5 & \bf 74.1 & \bf 81.5\\
\midrule
\GAR \emph{best query} & \it 78.9 & \it 85.4 & \it 90.3 & \it 93.1 & \it 95.5 & \it 97.1 & \it 78.6 & \it 85.2 & \it 90.9\\

\midrule
& \multicolumn{9}{c}{\textit{Transfer from TriviaQA}}  \\
\midrule
DPR$^\dagger$ & \bf 56.8 & \bf 71.4 & \bf 81.2 & 78.8 & 87.9 & \bf 93.7 & 51.2 & 62.7 & 74.6  \\ 
\GAR & 45.5 & 61.8 & 76.7 & 71.5 & 84.0 & 91.5 & 58.2 & 68.9 & 78.7\\
\ours-RI & 49.6 & 67.1 & 79.6 & 74.2 & 86.2 & 92.5 & 62.1 & 72.0 & 80.4\\
\ours-RD & 54.5 & 68.0 & 79.7 & \bf 79.8 & \bf 88.5 & 93.1 & \bf 64.9 & \bf 73.0 & \bf 80.5\\
\midrule
\GAR \emph{best query} & \it 78.4 & \it 84.6 & \it 89.3 & \it 92.5 & \it 95.2 & \it 96.8 & \it 79.1 & \it 85.9 & \it 91.8 \\

\bottomrule
\end{tabular}
\caption{Top-$k$ retrieval accuracy on the test sets of three datasets for cross-dataset generalization settings. 
\GAR \emph{best query} represents the performance upper bound we can achieve by selecting the best query according to the labels.
Numbers in bold are the best scores for each setting. $^\dagger$Results are provided by \citet{ram-etal-2022-learning}.
}
\label{tab:cross}
\end{table*}
\begin{table}[t!]
\small
\centering
\begin{tabular}{lccc}
\toprule
\bf Model     & \bf NQ & \bf TriviaQA \\
\midrule
\multicolumn{3}{c}{\textit{Top-100 passages as input to FiD}} \\
\midrule
DPR + Extractive  & 41.5  &  57.9  \\
RAG & 44.5 & 56.1 \\
DPR + FiD & 51.4 & 67.6 \\
\GAR + FiD & 50.6 & 70.0 \\
SEAL + FiD & 50.7 & -\\
\citet{liu2022query} + FiD & 51.7 & 70.8 \\
\midrule
\ours RI + FiD& 51.4 & 71.2 \\
\ours RD + FiD & \bf 52.1 & \bf 71.5 \\
\midrule
\multicolumn{3}{c}{\textit{Top-10 passages as input to FiD}} \\
\midrule
\GAR + FiD & 30.5 & 48.9 \\
\ours RI + FiD& 35.5 & 56.7  \\
\ours RD + FiD & \bf 39.6 & \bf 60.0 \\
\bottomrule
\end{tabular}
\caption{
    End-to-end QA exact-match scores on the test sets of NQ and TriviaQA. Numbers for prior work are cited from \citet{liu2022query}.
}
\label{tab:fid}
\end{table}

\subsection{Result: Cross-Dataset Generalization}
To better evaluate the robustness of these models for out-of-domain examples, we train our models only on NQ or TriviaQA, and then test them on WebQ, TREC, and EntityQs in a \emph{zero-shot} manner. The results are shown in Table~\ref{tab:cross}. We observe that when transferring from NQ or TriviaQA, DPR experiences a decline in performance compared to in-domain supervised training on WebQ.\footnote{The in-domain DPR performs poorly on TREC since it only has 1,125 training examples.} \GAR performs even worse than DPR on both WebQ and TREC. However, \GAR performs better than DPR on EntityQs, which is designed to challenge dense retrieval by including many rare entities. Here we also present the performance of \emph{\GAR best query}. We see that although \GAR transfers poorly on cross-domain datasets, it still has the ability to generate high-quality query expansions by random sampling. This provides an opportunity for \ours to improve performance.
After adopting \ours, we see that \ours-RI improves the performance of \GAR by 2-4 points for top-5/20 accuracy, and \ours-RD further boosts the performance of \GAR by 5-10 points for top-5/20 accuracy. Overall, \ours-RD outperforms DPR except when transferring from TriviaQA to WebQ.

These results suggest that query reranking is a general technique that can work well even on out-of-domain examples, showing that \emph{generating relevant contexts} (\GAR) is largely dependent on the domains, while \emph{judging which contexts may be more beneficial to retriever} is a more domain-agnostic skill.

\begin{table*}[h!]
\centering
\small
\begin{tabular}{lcccccc} 
\toprule
\multirow{2.6}{55pt}{\bf Model}
& \multicolumn{3}{c}{\bf Natural Questions~}  
& \multicolumn{3}{c}{\bf TriviaQA}       \\ 
\cmidrule(lr){2-4} \cmidrule(lr){5-7}
& \bf Top-5 & \bf Top-20 & \bf Top-100
& \bf Top-5 & \bf Top-20 & \bf Top-100   \\ 
\midrule
BM25 & 43.8 & 62.9 & 78.3 & 67.7 &
77.3 & 83.9  \\
\GAR & 60.8 & 73.9 & 84.7 & 71.8 & 79.5 & 85.3 \\
\GAR+ Passage Rerank (k=25/k=32) & 68.8 & 75.7 & 84.7 & 77.6 & 81.0 & 85.3 \\
\GAR+ Passage Rerank (k=100) & 71.7 & 80.2 & 84.7 & 79.2 & 83.3 & 85.3 \\
\midrule
\ours-RD & 69.3 & 78.6 & \bf 86.5 & 77.6 & 82.1 & \bf 86.4 \\
\ours-RD + Passage Rerank (k=100) & \bf 73.7 & \bf 82.1 & \bf 86.5 & \bf 80.6 & \bf 84.5 & \bf 86.4 \\

\bottomrule
\end{tabular}
\caption{Top-k retrieval accuracy (\%) on the Natural Questions and TriviaQA test sets for comparison of query reranking and passage reranking.}
\label{tab:rerank}
\end{table*}

\subsection{Result: End-to-end QA with FiD}

To fully understand whether \ours can benefit end-to-end QA systems, we further evaluate the exact match scores with Fusion-in-Decoder (FiD)~\citep{izacard-grave-2021-leveraging}, a generative reader model trained from T5-large~\citep{JMLR:v21:20-074}. We take the FiD models that were pre-trained on NQ/TriviaQA and directly test on our retrieval results without further fine-tuning. The exact match scores using the top-100 retrieved passages as input to FiD is shown at the top of Table~\ref{tab:fid}. We observe that \ours consistently outperforms previous work, including DPR, \GAR, SEAL, and \citet{liu2022query}, on both NQ and TriviaQA.
Although these gains may appear relatively small, however, this is primarily due to FiD's ability to take the top-100 retrieved passages as input and generate answers using cross-attention across all passages. Thus, even with low-ranked answer passages (say the answer is in the 99th passage), it is still possible that FiD could produce correct answers.

As there are many methods where relatively smaller context windows compared to FiD are used, especially when models are scaled up and cross-attention becomes much more expensive, improving retrieval accuracy for smaller $k$ may be beneficial.
For example, GPT-3~\citep{brown2020language} only has a context window size of 2048, which can only support 10-20 passages as input.
We explore this setting by selecting only the top-10 retrieved passages as input to FiD, and show the results at the bottom of Table~\ref{tab:fid}.
\ours achieve significant improvement over \GAR, roughly 10\% in exact match on both datasets, showing potential benefits for methods with limited context window size.

\section{Query Reranking vs Passage Reranking}
\label{sec:psg_rerank}

\ours shares similarities with passage reranking (PR). \ours reranks the queries before retrieving the passages, while PR reranks the retrieved list of passages after the retrieval process is completed. To better understand the relationship between \ours and PR, we implement a BERT-based passage reranker, following the method outlined in \citet{nogueira2019passage}, to rerank the retrieval results of \GAR. The implementation details can be found in Appendix~\ref{app:reranker}. From the experiments we aim to answer three questions: 1) Is \ours better than PR? 2) Are the contributions of \ours and PR complementary? Can their performance gains be aggregated if we apply both? 3) What are the extra advantages of \ours compared to PR?

\paragraph{Is \ours better than PR?} We focus on comparing \ours-RD with PR, as \ours-RI is limited by its input, being able to see only the short expanded queries. On the other hand, \ours-RD has access to the top-1 passage retrieved by each query candidate, providing it with the same level of information as PR. In Table~\ref{tab:rerank}, we first present the performance of PR when reranking the same number of passages as the average number of query candidates considered by \ours (25 for NQ; 32 for TriviaQA), which can be found in row 3. The result of \ours-RD (shown in row 5) is better than row 3, indicating that when considering the same amount of information as inputs, \ours-RD outperforms PR. However, when PR is able to rerank a larger number of passages, such as the top-100 passages shown in row 4, it achieves better performance than \ours-RD. This implies that \ours-RD is more effective when PR can only access to the same level of information.

\paragraph{Are \ours and PR complementary?} We found that the effects of \ours-RD and PR can be effectively combined for even better performance. When applying PR on the retrieval results of \ours-RD (shown in row 6), we see a significant improvement compared to both row 4 and row 5. This suggests that the contributions of \ours-RD and PR are complementary: \ours strengthens first-pass retrieval by selecting good queries, while PR re-scores all the retrieved passages and generates an entirely new order for these passages. The distinction between these two mechanisms makes improvements accumulative and leads to superior results. 

\paragraph{Extra advantages of \ours?}An advantage of \ours is that it improves retrieval results beyond the top-k passages. In row 4, the top-100 accuracy cannot be improved by PR as it reranks within the top-100 passages. In contrast, the improvements provided by \ours are not limited to top-100 passages. As long as \ours selects good query expansions, it can improve the whole list of retrieved passages; we can see \ours-RD improves the top-100 accuracy of \GAR from 84.7 to 86.5.

\begin{table}[t!]
\centering
\small
\begin{tabular}{lccc}
\toprule
\bf Model & \bf Top-5 & \bf Top-20 & \bf Top-100 \\ 
\midrule
\ours-RI & 63.2 & 76.4 & 85.9 \\
\ours-RI holdout & 63.6 & 76.3 & 86.0 \\
\midrule
\ours-RD & 69.3 & 78.6 & 86.5 \\
\ours-RD w/ DPR & 65.7 & 78.7 & 86.3 \\
\bottomrule
\end{tabular}
\caption{Top-k retrieval accuracy (\%) on NQ for comparison of the two different training example construction methods and for \ours with dense retrievers.}
\label{tab:variant}
\end{table}

\section{Discussions}

\subsection{Generating Training Examples with \GAR}
\label{sec:t0}
In Section~\ref{sec:construct}, we discussed two methods to construct training examples for \ours. In our main experiments, we used T0-3B to randomly sample diverse query expansions. An alternative method was also explored, where we trained $K=5$ different \GAR models separately on $(K-1)$ training subsets, then randomly sampled from the hold-out sets. The performance of this method, as shown in Table~\ref{tab:variant} (\ours-RI holdout), is slightly better than using T0-3B, but the difference is less than 0.5 points on Top-5/20/100 accuracy. Therefore, we continue to use T0-3B to generate training data in our main experiments as it eliminates the need to train $K$ different \GAR models separately.

\begin{table}[!t]
\small
\centering
\begin{tabular}{lccc}
\toprule
\bf Model     & \bf RI & \bf RD \\
\midrule
\ours N=50 & 63.16 & 69.34 \\
\ours N=30 & 63.02 & 68.86 \\
\ours N=20 & 62.96 & 68.67 \\
\ours N=10 & 62.60 & 67.62 \\
\ours N=5 & 62.44 & 66.32 \\
\midrule
GAR baseline (N=1) & \multicolumn{2}{c}{60.80} \\
\bottomrule
\end{tabular}
\caption{
    Top-5 accuracy on NQ with different $N$. $N$ stands for the maximum number of query expansions considered by the query reranker.
}
\label{tab:reduce}
\end{table}

\begin{figure*}[ht!]
    \centering
    \begin{subfigure}{\columnwidth}
        \includegraphics[width=\linewidth]{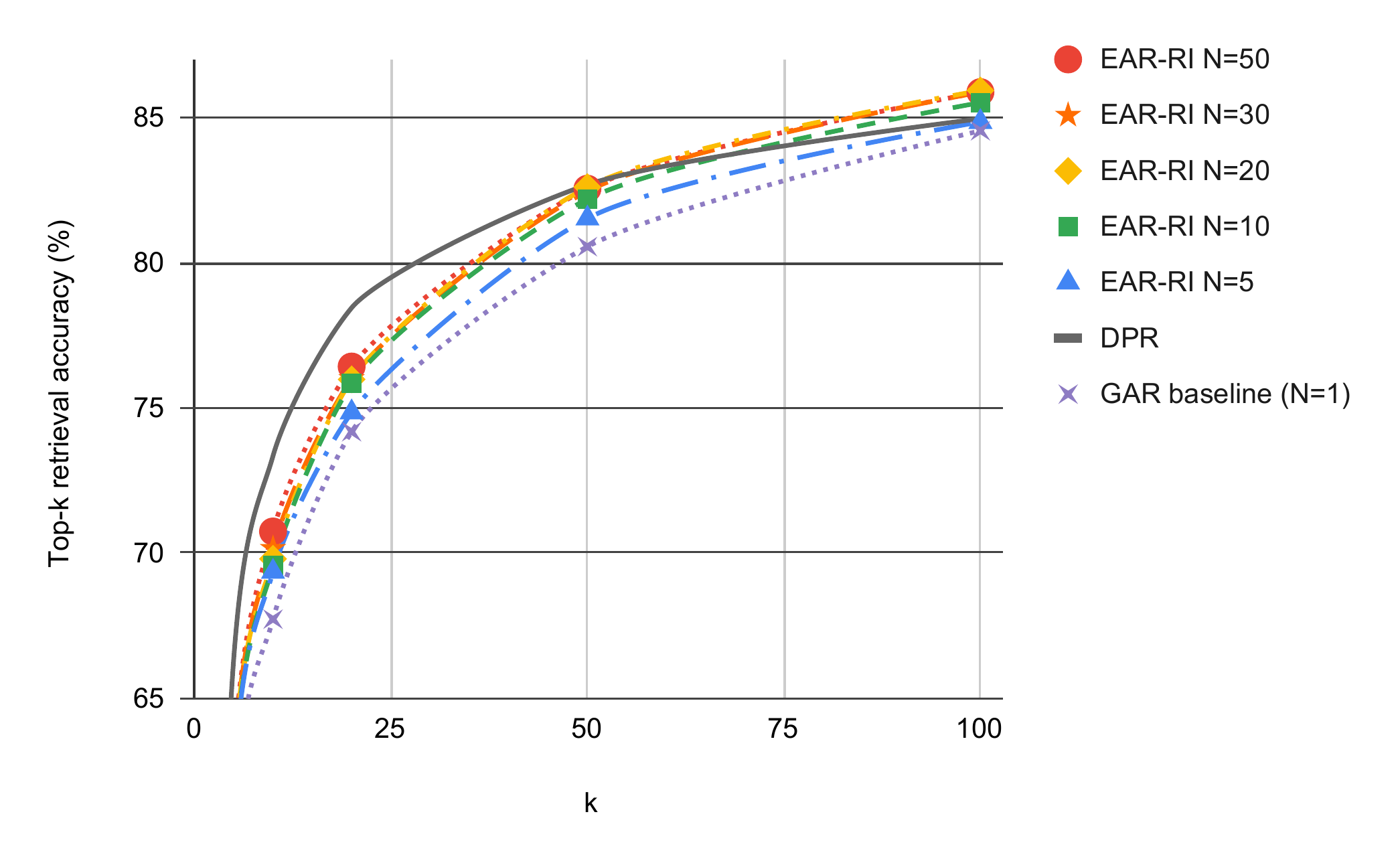}
        \caption{\ours-RI}
        \label{fig:reduce_ri}
        \end{subfigure}
    \begin{subfigure}{\columnwidth}
        \includegraphics[width=\linewidth]{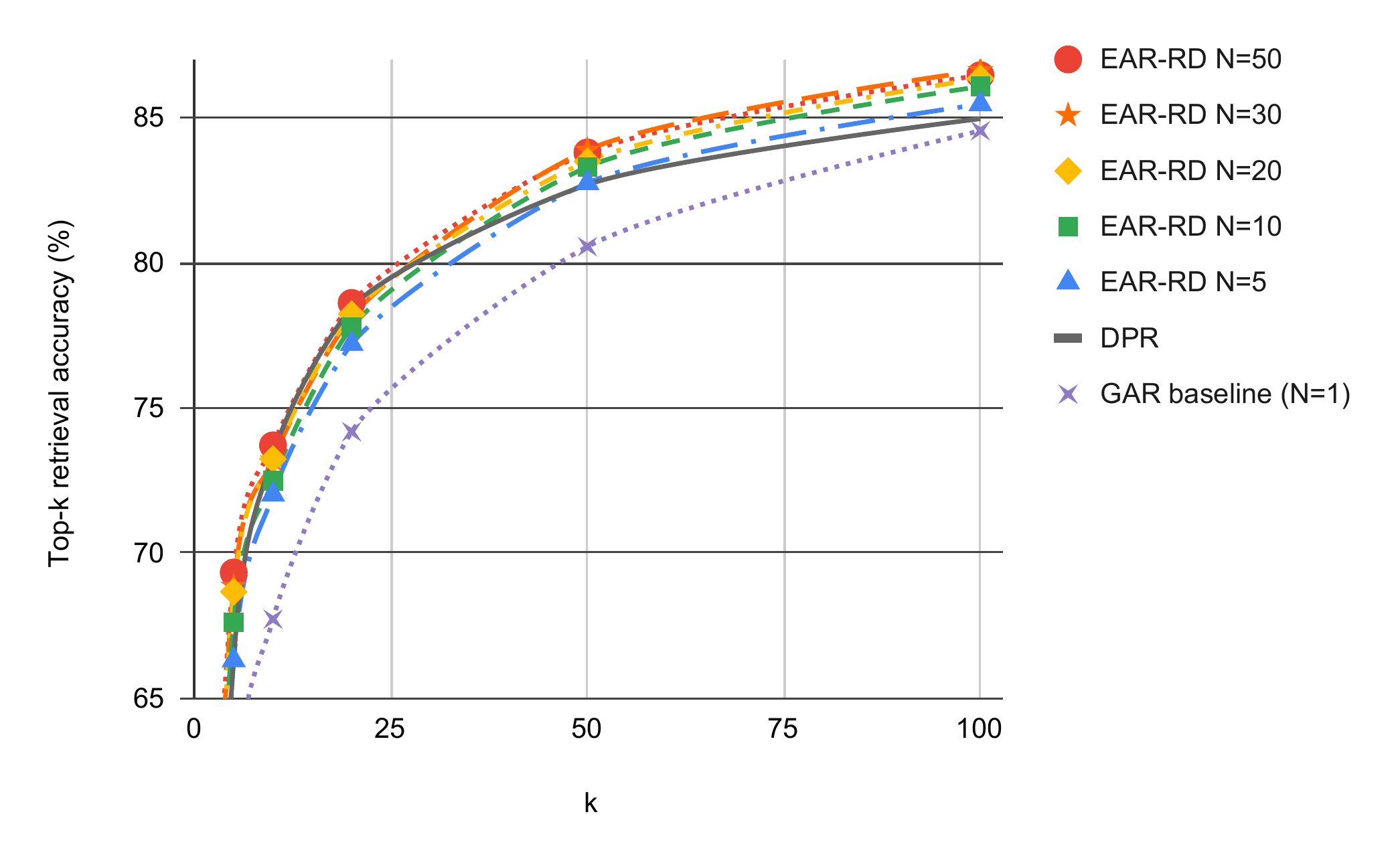}
        \caption{\ours-RD}
        \label{fig:reduce_rd}
    \end{subfigure}
    \caption{Top-k performance curves on NQ for \ours-RI and \ours-RD with a reduced candidate size $N$.}
    \label{fig:reduce}
\end{figure*}

\subsection{E\textsc{ar} with Dense Retrievers}
\ours is specifically optimized to work well with the BM25 retriever and hence its performance may be impacted when changing the retriever to DPR. As shown at the bottom of Table~\ref{tab:variant}, when coupled with DPR, the top-5 accuracy of \ours-RD decreases, while the top-20/100 accuracy remains relatively unchanged. This suggests that \ours is heavily reliant on the retriever, and thus changing the retriever negatively impacts its performance. Making \ours work with DPR would require retraining with DPR retrieval results and significantly more compute. We leave this direction for future work.

\subsection{Reducing the Query Candidate Size}

In our experiments, we generate 50 query expansions per question and then de-duplicate the repeated ones. However, we can also limit the maximum query expansions considered by our reranker to trade off between efficiency and performance. In Table~\ref{tab:reduce} we show the top-5 accuracy of lowering the maximum candidate size $N$ from 50 to 30/20/10/5. We observe that the performance drops gradually as $N$ decreases. However, we still see improvement over \GAR even when $N=5$, showing that \ours still works with a small candidate size. We also show the curves of the top-k accuracy in Figure~\ref{fig:reduce}, where we observe a big gap between DPR (solid line) and \GAR (dotted line with x mark). \ours-RI gradually eliminates the gap as $N$ increase, while \ours-RD even matches DPR for $k < 50$ and outperforms DPR for $k \geq 50$ with a small $N=5$. 

\section{Computational Cost and Latency}
\label{sec:latency}

We report the latency of DPR, \GAR, and \ours in Table~\ref{tab:latency}. Inference details can be found in Appendix~\ref{app:inference}. 
\paragraph{Dense Retrieval} We first generate DPR document embeddings on 4 GPUs for \mytilde{}3.5 hours on 21M documents. Standard indexing takes \mytilde{}10 minutes with a 64GB index size. Indexing with the more advanced Hierarchical Navigable Small World (HNSW)~\citep{malkov2018efficient} takes \mytilde{}5 hours and results in a huge index size of 142GB. For retrieval, standard indexing takes 22.3s per query, while the highly optimized HNSW can shorten it to 0.04s per query.

\begin{table}[t!]
\centering
\scriptsize
\begin{tabular}{l|p{0.08\linewidth}p{0.11\linewidth}p{0.07\linewidth}p{0.07\linewidth}|p{0.07\linewidth}|p{0.08\linewidth}} 
\toprule
\bf Model & \bf Build Index & \bf Query Expand & \bf Query Rerank & \bf Retri- eval
& \bf Index Size & \bf Top-5 (NQ)   \\ 
\midrule
DPR & 3.5hr & - & - & 22.4s & 64GB & 68.3 \\
+HNSW & 8.5hr & - & - & 0.04s & 142GB & 68.0 \\
\midrule
BM25 & 0.5hr & - & - & 0.15s & 2.4GB & 43.8 \\
\GAR & 0.5hr & 0.58s & - & 0.56s & 2.4GB & 60.8 \\
\midrule
\ours-RI & 0.5hr & 1.29s & 0.04s & 0.50s & 2.4GB & 63.2 \\
\ours-RD & 0.5hr & 1.29s & 0.84s & 0.54s & 2.4GB & 69.3 \\

\bottomrule
\end{tabular}
\caption{Latency per query for DPR/BM25/\GAR/\ours.}
\label{tab:latency}
\end{table}

\paragraph{Sparse Retrieval} For BM25 with Pyserini, indexing only takes 0.5 hours, with a very small index size of 2.4GB. Retrieval for BM25 takes 0.15s per query. For \GAR, it needs an extra 0.58s to generate the query expansions, and retrieval time is 0.56s. For \ours, it needs 1.29s to batch sample 50 query expansions. \ours-RI only takes 0.04s to rerank queries. \ours-RD needs extra time to retrieve the top-1 passages for each expansion, which takes an extra 0.70s, and then run the actual reranking process, taking 0.14s, giving a total of 0.84s for query reranking. For retrieval, the time needed for both \ours-RI and \ours-RD is similar to \GAR.

To conclude, \ours inherits the advantage of BM25: \emph{fast indexing time} and \emph{small index size}. This makes it possible to index large collections of documents in a relatively short amount of time, which is important for tasks where documents are frequently added or updated.
The main cost for \ours is the time for sampling query expansions. However, this can potentially be reduced by speed-up toolkits that optimize the inference time of transformers, such as FasterTransformer~\footnote{\url{https://github.com/nvidia/fastertransformer}} (3.8\mytilde{}13$\times$ speedup for decoding) or FastSeq~(\citealp{yan2021fastseq};  7.7$\times$ speedup for BART decoding). Moreover, we can leverage model distillation~\citep{shleifer2020pre} and quantization~\citep{li2022dq} for transformers. 
We leave these directions for future work.

\section{Related Work}
\paragraph{Query Expansion and Reformulation}
Traditionally, query expansion methods based on pseudo relevance feedback utilize relevant context without external resources to expand queries~\citep{rocchio71relevance,Jaleel2004UMassAT,10.1145/1835449.1835546,10.1145/3459637.3482124}. 
Recent studies attempt to reformulate queries using generative models, relying on external resources such as search sessions~\citep{10.1145/3397271.3401323} or conversational contexts~\citep{Lin2020QueryRU,vakulenko2021question}, or involve sample-inefficient reinforcement learning training~\citep{nogueira-cho-2017-task}. 
More recently, \GAR~\citep{mao-etal-2021-generation} explored the use PLMs for query expansion instead of external resources.
A concurrent study~\citep{liu2022query} generates multiple expansions with beam search and filters and fuses the results, but \ours is aware of the BM25 retriever and could select more promising query expansions and run fewer BM25 retrievals.

\paragraph{Retrieval for OpenQA}
Sparse retrieval with lexical features such as BM25 was first explored for OpenQA~\cite{chen-etal-2017-reading}. 
Dense retrieval methods were shown to outperform sparse methods~\cite{karpukhin-etal-2020-dense, 10.5555/3524938.3525306}, while requiring large amounts of annotated data and much more compute. 
Although powerful, dense retrievers often fall short in the scenarios of 1) requiring lexically exact matching for rare entities~\cite{sciavolino-etal-2021-simple} and 2) out-of-domain generalization~\citep{reddy2021towards}. For 1), \citet{luan-etal-2021-sparse} proposed a sparse-dense hybrid model, and \citet{chen2021salient} trained a dense retriever to imitate a sparse one. For 2), \citet{ram-etal-2022-learning} created a pre-training task for dense retrievers to improve zero-shot retrieval and out-of-domain generalization.
Another recent line of research explores passage reranking with PLMs to improve performance for both sparse and dense methods. 
\citet{nogueira2019passage} first explored BERT-based supervised rerankers for standard retrieval tasks and \citet{mao-etal-2021-reader} proposed reranking by reader predictions without any training.
\citet{sachan-etal-2022-improving} attempt to use an LLM directly as the reranker, but it requires huge amounts of computation at inference time and underperforms fine-tuned rerankers. 

\section{Conclusion}

We propose \ours, which couples \GAR and BM25 together with a query reranker to unlock the potential of sparse retrievers. \ours significantly outperforms DPR while inheriting the advantage of BM25: \emph{fast indexing time} and \emph{small index size} compared to the compute-heavy DPR. Cross-dataset evaluation also shows that \ours is very good at generalizing to out-of-domain examples. Furthermore, we demonstrate that contributions of \ours and passage reranking are complementary, and using both methods together leads to superior results. Overall, \ours is a promising alternative to existing dense retrieval models, providing a new way to achieve high performance with less computing resources.

\section*{Limitations}
First, as \ours largely relies on \GAR generators, the performance of the method is closely tied to the quality of the generator used. We have attempted to use large language models such as T0-3B without fine-tuning as a replacement for the \GAR generator during testing, but the performance becomes worse. The main reason is that the quality of query expansions generated by T0-3B is too diverse, which makes \ours has a higher chance to select from terrible expansions. In contrast, the output quality of \GAR is more stable. We may need a more complex mechanism that can exclude terrible query expansion if we want to directly use the query expansions generated by T0-3B during inference.
Second, \ours has demonstrated a strong generalization ability to out-of-domain data, but the method may still face challenges when transferring to other languages without any supervised QA data, which \GAR and \ours are trained on. Although challenging, we are still trying to train the \ours system without supervised QA data.

\section*{Ethics Statement}

In this research, we used publicly available datasets and we did not collect any personal information. 
Our method is designed to improve the performance of information retrieval systems, which can have a positive impact on various applications, such as search engines, QA systems, and other applications that rely on text retrieval. 
When deployed, however, our approach also poses the ethical risk typical of pre-trained language models, for instance, producing retrieval results that contain human biases which could potentially exacerbate discrimination.
Therefore, caution should be exercised before implementing our approach in real-world situations and thorough audit of training data and testing of model outputs should be conducted.

\section*{Acknowledgements}

We thank Yuning Mao for his helpful feedback. We thank Ori Ram for providing detailed results of the experiments from~\citet{ram-etal-2022-learning}.

\bibliography{anthology,custom}
\bibliographystyle{acl_natbib}

\newpage
\appendix

\section{Dataset Statistics}
\label{app:data}

We show the number of train/dev/test examples in each dataset in Table~\ref{tab:dataset_stats}.
\begin{table}[h!]
\small
\centering
\begin{tabular}{lrrr}
\toprule
\textbf{Dataset} & \textbf{Train} & \textbf{Dev} & \textbf{Test} \\
\midrule
Natural Questions & 58,880 & 8,757 & 3,610 \\
TriviaQA & 60,413 & 8,837 & 11,313 \\
WebQuestions & - & - & 2,032 \\
TREC & - & - & 694 \\
EntityQs & - & - & 22,075 \\
\bottomrule
\end{tabular}
\caption{Number of train/dev/test examples in each dataset.}
\label{tab:dataset_stats}
\end{table}

\section{Training Details}
\label{app:training}

For the training set, we use T0-3B~\footnote{\url{https://huggingface.co/bigscience/T0\_3B}} to randomly sample 50 query expansions per query.
For the dev set and test set, we use the three \GAR generators (answer/sentence/title), which are BART-large seq2seq models~\cite{lewis-etal-2020-bart} to generate 50 query expansions per query.
We use the DeBERTa V3 base model\footnote{\url{https://huggingface.co/microsoft/deberta-v3-base}}, which has 86M parameters that are the same as BERT-base~\citep{devlin-etal-2019-bert}, as \ours-RI and \ours-RD rerankers.
For the implementation of rerankers, we reference the implementation of SimCLS~\citep{liu-liu-2021-simcls}\footnote{\url{https://github.com/yixinL7/SimCLS}}, which also does reranking for sequences. We start from the code of SimCLS and change the loss function to our ranking loss $\mathcal{L}_{\text{Rank}}$.  
During training, we use the dev set generated from three \GAR generators to pick the best checkpoints, resulting in three different reranker models corresponding to the answer/sentence/title generators.

The ranges we search for our hyperparameters are shown in Table~\ref{tab:hyper}. Each training example in our dataset contains 50 sequences (generated by T0-3B). To prevent memory issues of GPU, we used gradient accumulation to simulate a batch size of 4 or 8, which effectively consists of 200 or 400 sequences, respectively.

The training time on a single NVIDIA V100 GPU is around 12 hours for \ours-RI and 1 to 2 days for \ours-RD. The best hyperparameters according to the dev set are shown in Table~\ref{tab:best}. However, in our experiments, the variance between different hyperparameters is actually quite small.

\begin{table}[h!]
\small
\centering
\begin{tabular}{lc}
\toprule
\textbf{Hyperparams} & \textbf{Range} \\
\midrule
\texttt{MAX\_RANK} & [101, 250] \\
Batch size & [4, 8] \\
Learning rate & [2e-3, 5e-3] \\
Epochs (\ours-RI) & 2 \\
Epochs (\ours-RD) & 3 \\
Max length (\ours-RI) & 64 \\
Max length (\ours-RD) & 256 \\
\bottomrule
\end{tabular}
\caption{The range for hyperparameter search. The definition of \texttt{MAX\_RANK} is shown in Section~\ref{sec:ri}.}
\label{tab:hyper}
\end{table}

\begin{table}[h!]
\small
\centering
\begin{tabular}{lccc}
\toprule
\textbf{Hyperparams} & \textbf{answer} & \textbf{sentence} & \textbf{title} \\
\midrule
\multicolumn{3}{l}{\textit{NQ}: \ours-RI } \\
\midrule
\texttt{MAX\_RANK} & 101 & 250 & 101 \\
Batch size & 8 & 8 & 4 \\
Learning rate & 5e-3 & 2e-3 & 2e-3  \\
\midrule
\multicolumn{3}{l}{\textit{NQ}: \ours-RD } \\
\midrule
\texttt{MAX\_RANK} & 101 & 101 & 250 \\
Batch size & 4 & 4 & 8\\
Learning rate & 2e-3 & 2e-3 & 2e-3 \\
\midrule
\multicolumn{3}{l}{\textit{TriviaQA}: \ours-RI } \\
\midrule
\texttt{MAX\_RANK} & 101 & 101 & 101 \\
Batch size & 8 & 8 & 8 \\
Learning rate & 2e-3 & 2e-3 & 2e-3 \\
\midrule
\multicolumn{3}{l}{\textit{TriviaQA}: \ours-RD } \\
\midrule
\texttt{MAX\_RANK} & 101 & 101 & 101 \\
Batch size & 8 & 8 & 8 \\
Learning rate & 2e-3 & 2e-3 & 2e-3 \\
\bottomrule
\end{tabular}
\caption{The best hyperparameters for NQ and TriviaQA dev sets.}
\label{tab:best}
\end{table}

\section{Passage Reranking}
\label{app:reranker}
For the implementation of a BERT-based passage reranker, we generally follow the setting of~\citep{nogueira2019passage} for training.
We separately fine-tuned two \texttt{bert-base-uncased} models on the NQ training set and the TriviaQA training set. 
Each pre-trained BERT model is fine-tuned for reranking using cross-entropy loss on the binary classification head on top of the hidden state corresponding to the [CLS] token.
We use the top-10 outputs of BM25 ran on the training sets as the training examples, which contains both positive and negative examples.
We fine-tune the models using 2 GPUs with mixed precision (fp16) with a batch size of 128 for 3 epochs.
AdamW~\citep{loshchilov2018decoupled} is used for optimization with a learning rate of 5e-5, linear warmup over the first 10k steps and linear decay afterwards, and a weight decay of $0.01$.

\section{Inference Details}
\label{app:inference}

For inference of \GAR retrieval results, we follow \GAR to retrieve with three queries generated by three context generators (answer/sentence/title), and then fuse the three retrieved passages lists in the order of sentence, answer, title. In other words, given the three retrieved lists of passages: $(a_1, a_2, ..., a_{100})$, $(s_1, s_2, ..., s_{100})$, $(t_1, t_2, ..., t_{100})$, we fuse the results as $(s_1, a_1, t_1, s_2, a_2, t_2, ..., s_{33}, a_{33}, t_{33}, s_{34})$. We skip all the duplicated passages that exist twice during the fusion process.

For \ours, we use the same pipeline of \GAR, while the only difference is that instead of greedy decoding, now the three generators of \GAR can do random sampling, and three different query rerankers (answer/sentence/title) are applied to select the best queries. After that, the pipeline to obtain retrieval results is exactly the same as \GAR.

To fairly compare the latency of these methods, we run the 3610 queries in NQ test set one-by-one without batching (batch size = 1) and compute the average the latency per query, where document encoding, query expansion and reranking are run on NVIDIA RTX A5000 GPUs and indexing and retrieval are run on fifty Intel Xeon Gold 5318Y CPUs @ 2.10GHz, for both FAISS~\citep{johnson2019billion} (DPR) and Pyserini~\citep{Lin_etal_SIGIR2021_Pyserini} (BM25). 

\paragraph{DPR document indexing}
We used 4 GPUs to encode 21M Wikipedia passages in parallel with mixed precision (fp16), which takes around 3.5 hours.
\paragraph{\GAR and \ours}
For inference of \GAR and \ours, answer/sentence/title generators/rerankers are run in parallel on three GPUs. 
\paragraph{FiD}
We take the public checkpoints of FiD\footnote{\url{https://github.com/facebookresearch/FiD}}, which are trained from T5-Large~\citep{raffel2020exploring} with NQ/TriviaQA, to directly evaluate the end-to-end QA performance. 

\section{Qualitative Study}
\label{app:qual}

\begin{table}[h!]
\small
\centering
\begin{tabular}{lccc}
\toprule
\textbf{Model} & \textbf{answer} & \textbf{sentence} & \textbf{title} \\
\midrule
Original Query & \multicolumn{3}{c}{9.2} \\
\midrule
\GAR & 13.3 & 38.8 & 32.3 \\
\ours-RI & 13.1 & 36.2 & 29.3  \\
\ours-RD & 13.2 & 38.2 & 28.8  \\
\bottomrule
\end{tabular}
\caption{Lengths of the expanded queries in words for different methods on NQ test set.}
\label{tab:lengths}
\end{table}

\begin{table*}[h!]
\small
\centering
\begin{tabular}{llcc}
\toprule
\bf Model     & \bf Query [\textit{Answer = May 18, 2018}] & \bf Answer Rank \\
\midrule
BM25  & \it When is the next \textbf{Deadpool} movie being \textbf{released}?  &  77  \\
\midrule
\multirow{2}{*}{\GAR} & \it When is the next \textbf{Deadpool} movie being \textbf{released}? \color{blue}{Miller Brianna Hildebrand Jack}   & \multirow{2}{*}{>100} \\
 & \it \color{blue}{Kesy Music by Tyler Bates Cinematography Jonathan Sela Edited by Dirk Westervelt...} &  \\
\midrule
\multirow{2}{*}{\ours-RI} & \it When is the next \textbf{Deadpool} movie being \textbf{released}? \color{blue}{Deadpool 2 is \colorbox{green!50}{scheduled} to be} & \multirow{2}{*}{1} \\
 & \it \color{blue}{released on May 26, \colorbox{green!50}{2018}, with \colorbox{green!50}{Leitch} directing.} &  \\
\midrule
\multirow{2}{*}{\ours-RD} & \it When is the next \textbf{Deadpool} movie being \textbf{released}? \color{blue}{The film is \colorbox{green!50}{scheduled} to be released} & \multirow{2}{*}{1} \\
 & \it \color{blue}{on March 7, \colorbox{green!50}{2018}, \colorbox{green!50}{in the United States}.} & \\
\midrule
\multicolumn{3}{c}{\bf Answer Passage} \\
\midrule
\multicolumn{3}{c}{\it "\textbf{Deadpool} 2" premiered at Leicester Square in London on May 10, 2018. It was \textbf{released} } \\
\multicolumn{3}{c}{\it \colorbox{green!50}{in the United States} on May 18, \colorbox{green!50}{2018}, having been previously \colorbox{green!50}{scheduled} for release } \\
\multicolumn{3}{c}{\it on June 1 of that year. \colorbox{green!50}{Leitch}\'s initial cut of the film was around two hours and twelve minutes, ...} \\
\bottomrule
\vspace{10pt}
\end{tabular}

\begin{tabular}{llcc}
\toprule
\bf Model     & \bf Query [\textit{Answer = Natasha Bharadwaj, Aman Gandotra}] & \bf Answer Rank \\
\midrule
BM25  & \it Who has won \textbf{India's next} super \textbf{star}?  &  96  \\
\midrule
\multirow{2}{*}{\GAR} & \it Who has won \textbf{India's next} super \textbf{star}? \color{blue}{The \colorbox{green!50}{winner} of the competition is 18 year-old}    & \multirow{2}{*}{>100} \\
 & \it \color{blue}{Mahesh Manjrekar from Mumbai.} &  \\
\midrule
\multirow{2}{*}{\ours-RI} & \it Who has won \textbf{India's next} super \textbf{star}? \color{blue}{The \colorbox{green!50}{winner} of the \colorbox{green!50}{Superstar} \colorbox{green!50}{Season} \colorbox{green!50}{2018}} & \multirow{2}{*}{1} \\
 & \it \color{blue}{is Siddharth Shukla.} &  \\
\midrule
\multirow{2}{*}{\ours-RD} & \it Who has won \textbf{India's next} super \textbf{star}? \color{blue}{The \colorbox{green!50}{winner} of the \colorbox{green!50}{Superstar} \colorbox{green!50}{Season} \colorbox{green!50}{2018}} & \multirow{2}{*}{1} \\
 & \it \color{blue}{is Siddharth Shukla.} &  \\
\midrule
\multicolumn{3}{c}{\bf Answer Passage} \\
\midrule
\multicolumn{3}{c}{\it \textbf{India's Next} \colorbox{green!50}{Superstars} (INS) is an Indian talent-search reality TV show, which premiered } \\
\multicolumn{3}{c}{\it on \textbf{Star} Plus and is streamed on Hotstar. Karan Johar and Rohit Shetty are the judges for the show.} \\
\multicolumn{3}{c}{\it Aman Gandotra and Natasha Bharadwaj were declared \colorbox{green!50}{winners} of the \colorbox{green!50}{2018} \colorbox{green!50}{season}...} \\
\bottomrule
\vspace{10pt}
\end{tabular}

\begin{tabular}{llcc}
\toprule
\bf Model     & \bf Query [\textit{Answer = Anthropomorphism, Pathetic fallacy, Hamartia, Personification}] & \bf Answer Rank \\
\midrule
BM25  & \it \textbf{Method} used by a \textbf{writer} to develop a character? &  92  \\
\midrule
\multirow{2}{*}{\GAR} & \it \textbf{Method} used by a \textbf{writer} to develop a character?  \color{blue}{Developing a character is a technique }   & \multirow{2}{*}{>100} \\
 & \it \color{blue}{employed by writers in the creation of a narrative.} &  \\
\midrule
\multirow{3}{*}{\ours-RI} & \it \textbf{Method} used by a \textbf{writer} to develop a character?  \color{blue}{Developing a character is the primary } & \multirow{3}{*}{>100} \\
 & \it \color{blue}{method employed by writers in the creation of a fictional character.} &  \\
\midrule
\multirow{3}{*}{\ours-RD} & \it \textbf{Method} used by a \textbf{writer} to develop a character?  \color{blue}{Developing a character is a technique } & \multirow{3}{*}{>100} \\
 & \it \color{blue}{employed by writers in terms of establishing a persona and building a relationship} &  \\
 & \it \color{blue}{between the reader and the character.} &  \\
\midrule
\multicolumn{3}{c}{\bf Answer Passage} \\
\midrule
\multicolumn{3}{c}{\it The intensive journal \textbf{method} is a psychotherapeutic technique largely developed in 1966 at Drew } \\
\multicolumn{3}{c}{\it University and popularized by Ira Progoff (1921-1998). It consists of a series of writing exercises} \\
\multicolumn{3}{c}{\it using loose leaf notebook paper in a simple ring binder, divided into sections to helping accessing } \\
\multicolumn{3}{c}{\it various areas of the \textbf{writer}'s life. These include a dialogue section for the personification of things, } \\
\multicolumn{3}{c}{\it a "depth dimension" to aid in accessing the subconscious and other places for ....} \\
\bottomrule
\end{tabular}
\caption{
    Examples that show the difference between BM25/\GAR/\ours-RI/\ours-RD. {\color{blue}{Words in blue}} are query expansions generated by \GAR. \textbf{Bold words} are useful keywords from the original query. \colorbox{green!50}{Words highlighted in green} are useful keywords generated by \GAR. \textbf{Answer Rank} shows the ranking of the answer passage in the retrieval results.
}
\label{tab:qualitative}
\vspace{-5pt}
\end{table*}

In this section, we aim to investigate the differences between queries generated by \GAR and \ours. We first look at the lengths of the expanded queries for \GAR, \ours-RI, \ours-RD. In general, the lengths of queries from \ours are slightly shorter than that of \GAR, but the trends are not very obvious. Thus, we conduct a qualitative study to see what is the difference between these queries.

As shown in Table~\ref{tab:qualitative}, we provide three examples to demonstrate how our method \ours works. In the first example, the initial query only includes two keywords, "Deadpool" and "released," that can match the answer passage. As a result, the BM25 algorithm is unable to retrieve the correct passage within the top results until the 77th passage. The greedy decoding output for \GAR also fails to retrieve the correct passage, as it includes many irrelevant name entities. However, both \ours-RI and \ours-RD are able to select useful outputs from \GAR, which contain keywords such as ``scheduled,'' ``2018,'' ``Leitch,'' and ``in the United States.'' Although none of these keywords contains the real answer \emph{May 18, 2018}, these keywords already provide enough lexical overlap with the answer passage, allowing BM25 to correctly retrieve the answer passage in the top-1 result.

For the second example, the original query only contains three keywords ``India's,'' ``next,'' and ``star'' that can match the answer passage, so BM25 with the original query cannot retrieve the correct passage within the top retrieved results until the 96th passage.
For \GAR, the greedy decoding output for \GAR is also not effective, as it is a misleading answer and only includes one useful keyword ``winner'' and thus cannot retrieve the correct passage within the top-100 results.
For \ours-RI and \ours-RD, they are able to select a sentence that, while not containing the correct answer ``Natasha Bharadwaj'' or ``Aman Gandotra,'' does include useful keywords such as ``winner,'' ``Superstar,'' ``Season,'' and ``2018.'' These keywords provide enough lexical overlap with the answer passage, allowing \ours-RI and \ours-RD to correctly retrieve the answer passage in the top-1 result.

The third example presents a challenging scenario. The initial query only includes two common keywords, ``method'' and ``writer,'' which is difficult to match the answer passage. While BM25 is able to correctly retrieve the answer at the 92nd passage, the generated query expansions are not helpful and instead are misleading, resulting in \GAR and \ours-RI/\ours-RD all unable to retrieve the correct passage within the top-100 results due to the distracting query expansions. This example illustrates the importance of the \GAR generators. If all of the generated query expansions are not useful, \ours is unable to improve the results.

\end{document}